\documentclass[sigconf]{acmart}

\usepackage[ruled,linesnumbered]{algorithm2e}
\usepackage{multirow}

\AtBeginDocument{%
  \providecommand\BibTeX{{%
    \normalfont B\kern-0.5em{\scshape i\kern-0.25em b}\kern-0.8em\TeX}}}

\setcopyright{acmlicensed}
\copyrightyear{2024}
\acmYear{2024}
\acmDOI{XXXXXXX.XXXXXXX}

\acmConference[xxxx]{xxxx}{2024}{}
\acmISBN{978-1-XXXX-XXXX-X/XX/XX}

\begin{document}

\title{Improving the Transferability of 3D Point Cloud Attack via Spectral-aware Admix and Optimization Designs}

\author{Shiyu Hu, Daizong Liu, Wei Hu}
\affiliation{%
  \institution{Wangxuan Institute of Computer Technology, Peking University}
  \city{Beijing}
  \country{China}
}


\begin{abstract}
Deep learning models for point clouds have shown to be vulnerable to adversarial attacks, which have received increasing attention in various safety-critical applications such as autonomous driving, robotics, and surveillance. 
Existing 3D attackers generally designs various attack strategies in the white-box setting, requiring the prior knowledge of 3D model details. However, real-world 3D applications are in the black-box setting, where we can only acquire the outputs of the target classifier.
Although few recent works try to explore the black-box attack, they still achieve limited attack success rates (ASR).
To alleviate this issue, this paper focuses on attacking the 3D models in a transfer-based black-box setting, where we first carefully design adversarial examples in a white-box surrogate model and then transfer them to attack other black-box victim models.
Specifically, we propose a novel Spectral-aware Admix with Augmented Optimization method (SAAO) to improve the adversarial transferability.
In particular, since traditional Admix strategy are deployed in the 2D domain that adds pixel-wise images for perturbing, we can not directly follow it to merge point clouds in coordinate domain as it will destroy the geometric shapes. Therefore, we design spectral-aware fusion that performs Graph Fourier Transform (GFT) to get spectral features of the point clouds and add them in the spectral domain. Afterwards, we run a few steps with spectral-aware weighted Admix to select better optimization paths as well as to adjust corresponding learning weights. At last, we run more steps to generate adversarial spectral feature along the optimization path and perform Inverse-GFT on the adversarial spectral feature to obtain the adversarial example in the data domain.
In this way, the Admix-optimized adversarial examples can achieve high transfer ASR.
Experiments show that our SAAO achieves better transferability compared to existing 3D attack methods.
\end{abstract}





\maketitle

\section{Introduction}
\begin{figure}
     \centering
     \includegraphics[width=0.5\textwidth]{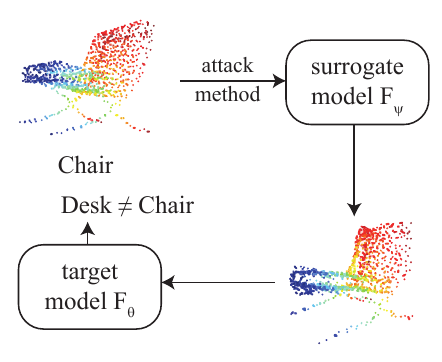}
     \caption{Transferbility is the ability of a attack method using adversarial examples generated on the known surrogate model to fool the unknown target model.}
     \label{fig:introduction}
 \end{figure}
Deep neural networks have shown to be vulnerable to adversarial examples \cite{goodfellow2014explaining,szegedy2013intriguing,liu2024a,liu2024pandora}, which add visually indistinguishable perturbations to network inputs but lead to incorrect prediction results. 
Significant progress has been made in adversarial attacks on 2D images, where many methods \cite{dong2018boosting,madry2017towards,kurakin2016adversarial,tu2019autozoom} learn to add imperceptible pixel-wise noise.
Nevertheless, adversarial attacks on 3D depth or point cloud data are still relatively underexplored. Different from images, point clouds are discrete representations of 3D objects or scenes, receiving increasing attention in various 3D applications such as autonomous driving \cite{chen2017multi,huang2023dense,liu2024survey,liu2024joint,huang2024advancing,liu2024cross} and medical data analysis \cite{singh20203d}. Similarly to their 2D counterparts \cite{liu2020saanet,liu2022few,zhu2023rethinking,liu2023hypotheses,liu2024transform,liu2023conditional,liu2023filling,liu2024unsupervised,fang2023you,fang2024your,tang2022few,liu2023spts,tang2022you,tang2022optimal,tang2024textsquare,feng2023unidoc,feng2023docpedia,zhao2024multi,zhao2024harmonizing,wang2024pargobridgingvisionlanguagepartial,tang2023character}, deep learning models trained for point clouds are often vulnerable to adversarial perturbations, increasing the risk in safety-critical 3D applications.

Most existing 3D point cloud attack methods \cite{xiang2019generating,wicker2019robustness,zhang2019adversarial,zheng2019pointcloud,tsai2020robust,zhao2020isometry,zhou2020lg,hamdi2020advpc,liu2023point,liu2021spatiotemporal,tao20233dhacker,liu2023robust,liu2024explicitly,cai2024frequency,yang2025hiding,liu2024hard} generally adapt the existing 2D adversarial attacks into the 3D scenario.
They either follow a point addition/dropping framework that identifies critical points from point clouds and modifies (add or delete) them to distort the most representative features, or follow the C\&W framework \cite{goodfellow2014explaining} to learn to perturb the Euclidean.
While the methods have demonstrated high attack success rates (ASR), they primarily operate within a white-box framework, wherein attackers possess complete knowledge of victim models, including network structure and weights. However, this approach diminishes practicality as real-world 3D applications typically withhold model details from users. Recent efforts \cite{huang2022shape,liu2021imperceptible} have aimed to address 3D attacks in the black-box setting, devoid of prior model knowledge. Nevertheless, these approaches rely on geometric distance losses or additional shape information to implicitly confine perturbations, often resulting in distorted 3D object shapes with irregular surfaces or outliers, easily detectable by humans. Furthermore, these attacks often suffer from overfitting to the target network and lose their malicious intent when transferred to attack a different victim model.

Considering that transferability is a critical characteristic for evaluating the robustness of attack methods, in this paper, we focus on addressing the challenging transfer-based black-box attack.
Previous transfer-based attack methods \cite{hamdi2020advpc,zhang2024improving} solely focus on designing complicated adversarial architecture or complex adversarial transformation to improve the transferability of adversarial examples, failing to investigate the models' sensitivity and decision boundaries in depth. To this end, we put more effort into the optimization design during the adversarial example generation process and try to develop transferable samples against the model decision boundary.
Luckily, Admix algorithm ~\cite{Admix} is a traditional 2D strategy to enhance the transferability of perturbed images by updating adversarial examples iteratively with averaged gradients of mixed images and it steers adversarial examples to decision boundaries of other categories by selecting examples from different categories to mix with the current adversarial examples. However, we can’t simply do linear addition on point clouds, because unlike 2D images, 3D point clouds are unordered and linear addition on the spatial position may deform the geometry feature
of point clouds. Therefore, we have to get an ordered feature of point cloud to mix. Considering that graph spectral domain is able to quantify the ordered features of point cloud\cite{GFT, kipf2017semisupervised}, we propose to add the point clouds in the spectral domain and utilize the inverse transformation to decode the spectral features back to point-based object in the data domain.
In this way, we can develop spectral-aware Admix algorithm with further optimization strategies for improving the transferability of point cloud attack.

Specifically, we propose a novel Spectral-aware Admix with Augmented Optimization (SAAO) method to improve the adversarial transferability. In the SAAO, on the one side, we first utilize Graph Fourier Transform to get spectral features of the point clouds, on which we perform spectral-aware Admix to maintain the Geometry features of the point clouds. In the adjusted Admix, we linearly sample the point clouds from the target point cloud to the mix point cloud, which is called the augmentation path. We use learnable mix weights in the sample space to make better use of the spectral feature we extracted before and we use fixed weights in the spectral space to remain the Geometry structure and the attack imperceptibility. On the other side, we use a few steps to evaluate the performance of the augmentation paths approximately and use the adjusted Admix again on the selected paths. Finally, we use inverse Graph Fourier Transform to get the adversarial point cloud. Enhancing by this spectral-aware Admix with augmentation-based optimization strategy, the adversarial samples generated by SAAO has much higher transferability.

Overall, our contributions can be summarized into three aspects:
\begin{itemize}
    \item We make the attempt to improve the transferability of black-box attacks on 3D point clouds. Different from previous transfer-based attacks that designs complicated attack architecture, we delve into devise more efficient optimization process by fooling the decision boundary of 3D models.
    \item To improve the adversarial transferability, we devise a spectral-aware Admix algorithm to handle the unordered point cloud data. Augmentation-path strategy is also introduced to assist the optimization process for improving the robustness of adversarial samples.
    \item Extensive experiments demonstrate that our attack method achieves the highest transfer attack success rate while exhibiting competitive imperceptibility.
\end{itemize}

\section{Related Work}
\noindent \textbf{Point Cloud Learning.}
3D point cloud learning is an emerging topic in recent years, which has attracted many research interests. 
Among the latest popular 3D applications \cite{xu2020grid,weng2022context,yang2019learning,zhu2022vpfnet}, 3D object classification \cite{su2015multi,yu2018multi,ma2018learning,qiu2021geometric} is the most fundamental yet important point-based task, which requires the model to capture the representative knowledge including the local contexts of 3D object details and the global context of the geometric topology.
To learn the point cloud representation, early works \cite{shao2017attribute,de2017transform,hu2021overview} follow the optimization-based framework that utilizes transforms to build upon prior knowledge and characteristics of geometric data, which often need to strike a balance between structure and details.
Inspired by the significant advances in deep learning methods, recent works tend to develop large-scale 3D models to learn and distinguish the geometric features of point clouds, and can be grouped into two categories:
(1) Spatial methods: these methods \cite{simonovsky2017dynamic,shen2018mining,wang2019dynamic,xu2020grid} define operations in the spatial domain to capture the geometric contexts. As the pioneer works, DeepSets \cite{zaheer2017deep} and PointNet \cite{qi2017pointnet} are proposed to first extract the point-wise features and then utilize average or max pooling to learn the global information.
To capture more fine geometric details, PointNet++ \cite{qi2017pointnet++} and other extensive works \cite{duan2019structural,liu2019densepoint,yang2019modeling} are further proposed to utilize attention mechanism to learn the local geometric structure.
(2) Spectral methods: unlike the spatial methods that utilize general convolution implemented through MLP or pooling over spatial neighbors, the spectral methods define convolutions as spectral filtering. These works \cite{wang2019dynamic,atzmon2018point,thomas2019kpconv,liu2019relation,gao2020graphter} generally design spectral convolutions with eigenvectors of the graph Laplacian matrix to learn the point-to-point relations or approximate with the Chebyshev polynomials to reduce the computation complexity \cite{te2018rgcnn}.

\noindent \textbf{Adversarial Attacks on 3D Point Clouds.}
Many works \cite{xiang2019generating,wicker2019robustness,zhang2019adversarial,zheng2019pointcloud,tsai2020robust,zhao2020isometry,zhou2020lg,hamdi2020advpc,huang2022shape} investigate the vulnerability of 3D point clouds from the perspective of adversarial attack.
Early works \cite{xiang2019generating,zhang2019adversarial,wicker2019robustness,zheng2019pointcloud} simply modify a few points in the point cloud to achieve the attack. They either add limited synthesized points or drop the critical points based on the characteristics of 3D models.
Although they can achieve 100\% attack success rate, they easily lead to the outlier problem.
To alleviate this limitation, recent works propose to add perturbations to the whole point cloud.
These methods \cite{wen2020geometry,tsai2020robust,hamdi2020advpc,liu2019extending,ma2020efficient,zhang2019defense,liu2021imperceptible} learn to perturb the Euclidean coordinates of each point by utilizing the C\&W framework \cite{carlini2017towards} to shift the points to fool the 3D models. To preserve the original 3D object shape, they also utilize both Chamfer and Hausdorff distances as the constraint to keep the point cloud geometry.
However, they still fail to capture the geometric characteristics of the point cloud for preserving the geometry-dependent object shape.
Huang \textit{et al.} \cite{huang2022shape} consider geometric smoothness by designing pixel-wise perturbations to pull outliers back to the surface, which may fail to preserve sharp object contours since it is hard to determine the local surface for points with fast variation.
In addition, there are also a few works \cite{lee2020shapeadv,li2021pointba,xiang2021backdoor} that attack point clouds in the feature space or utilize the backdoor attack.

\section{Methodology}

\subsection{Notations and Problem Definition}
\noindent \textbf{3D point cloud attack.}
A point cloud consists of an unordered set of points $ P = \{p_i\}_{i=1}^n \in \mathbb{R}^{n \times 3}$ sampled from a 3D object or scene, where each point $p_i \in \mathbb{R}^3$ is a vector that contains the coordinates $(x, y, z)$ of point $i$, and $n$ is the number of points. In this paper, we focus on attacking the point cloud classification task, where $n$ is set to 1024 and each point cloud $P$ has ground-truth label $y_{true}$. The target classification model $f(\cdot)$ takes a point cloud $P$ as input and predicts a vector of confidence scores $f(P) \in \mathbb{R}^c$. The predicted label is $y = F(P) = argmax_{i\in Y} f(P)_i \in Y$, $Y = \{1, 2, 3, \cdots, c\}$, representing the class of the original 3D object underlying the point cloud, where $c$ is the
number of classes. We generate the slightly-perturbed adversarial sample $P^{adv} = P + \Delta$ by adding perturbation $\Delta$ on it. If $F(P^{adv})\neq y_{true}$, we call it a successful attack, and we evaluate the efficiency of the generating method by calculating the attack successful rate (ASR) and the perturbation size.

\noindent \textbf{Improving the transferability of 3D point cloud attack.}
Transferability is a critical characteristic for evaluating the robustness of attack methods. 
Most existing 3D attackers fail to consider this characteristic and solely focus on attacking the white-box models.
However, real-world applications are in the black-box setting, where we can only acquire the outputs of the target classifier $F_\theta$ with parameter $\theta$. Therefore, it is hard to utilize gradient optimization for generating adversarial samples on them.
To this end, it is natural to attack the target classifier $F_\theta$ with adversarial examples generated on another surrogate classifier $F_\psi$ with known parameter $\psi$. As long as both classifiers have good abilities to classify the point clouds, they should have similar decision boundaries, making it possible to adjust the adversarial examples with gradients of $F_\psi$ for attacking similar $F_\theta$. This ability to fool the target model is called transferability. 
In this paper, we focus on attacking the black-box model by using the transfer attack.
Improving the attack transferability is to improve the efficiency (or ASR) of generating adversarial samples using surrogate models to make sure $F_\theta(P^{adv})\neq y_{true}$, no matter what the target classifier $F_\theta$ is.

\subsection{Overview of Our Attack}

Pitcure 1 shows our attack baseline. We first perform GFT to get spectral features of the point clouds and for each point cloud in the batch, we choose some candidate point clouds with $y_i \neq y_{true}$, not needing their label to be same. Afterwards, we run a few steps with our spectral-aware weighted Admix to select better augmentation paths by cosine similarity, as well as to adjust our learning weights. Next, we run more steps to generate adversarial spectral feature along the augmentation path. Finally, we perform IGFT on the adversarial spectral feature to get the adversarial example.

\subsection{Our Attack Baseline using Spectral-aware Admix Algorithm}

\begin{figure*}
     \centering
     \includegraphics[width=\textwidth]{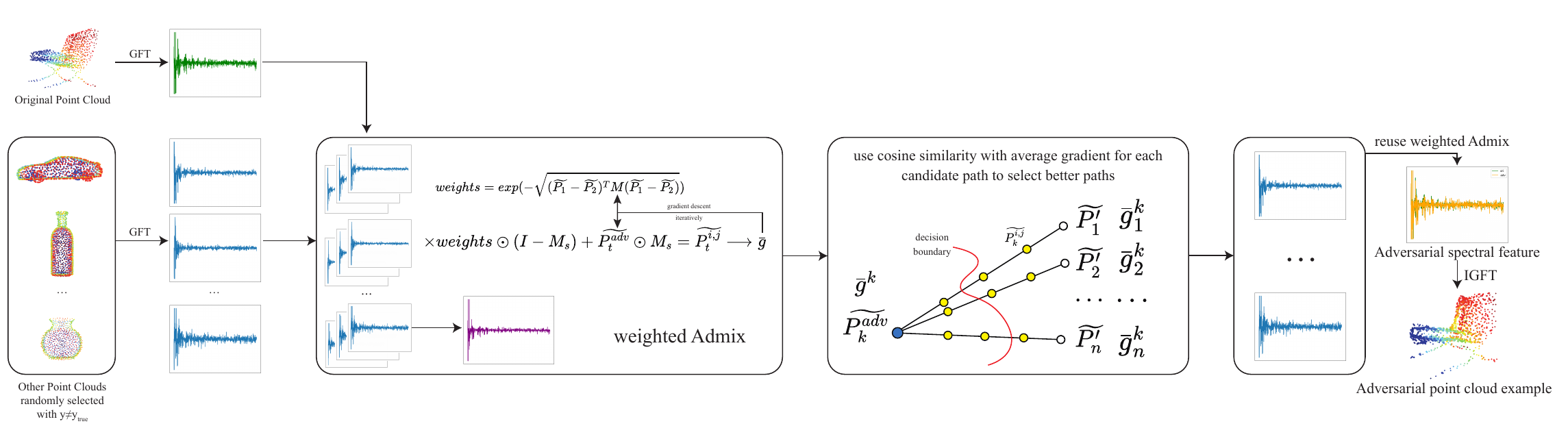}
     \caption{Our Attack baseline using weighted Admix and path selection.}
     \label{fig:baseline}
 \end{figure*}

\noindent \textbf{Preliminary of Admix.}
Admix~\cite{Admix} is proven to be a successful way of generating adversarial examples in 2D image classification tasks. It is designed to enhance the transferability of perturbed images by updating adversarial examples iteratively with averaged gradients of mixed images and it steers adversarial examples to decision boundaries of other categories by selecting examples from different categories to mix with the current adversarial example:
\begin{equation}
  \bar{g}_{t+1} = \frac{1}{m_1 \cdot m_2}\sum_{x' \in X'}^{} \sum_{i=0}^{m_1 - 1}\nabla _{x_t^{adv}}J(\gamma_i\cdot(x_t^{adv}+\eta_i \cdot x'),y;\theta),
\end{equation}
here, since images are ordered with fixed size, Admix can update the adversarial sample $x^{adv}$ of original images $x$ with averaged gradient $g$ of $m_1$ scaled copies of original image mixed with $m_2$ images $x' \in X'$ from other categories. $J$ is the target classifier with parameter $\theta$, $\gamma_i$ and $\eta$ are mixing weights, $t$ is the running step. It inspires us that we can boost transferability on point cloud attack task by mixing the point clouds properly.

\noindent \textbf{Spectral-aware Admix for 3D Domain.} Admix is successful in 2D tasks because it can directly linearly add two images into the intermediate sample for optimization. However, we can't simply do linear addition on point clouds, because unlike 2D images, 3D point clouds are unordered as we mentioned above, and linear addition on the spatial position may deform the geometry feature of point clouds. Therefore, we have to get an ordered feature of point cloud to mix. 
Considering that graph spectral domain is able to quantify the ordered features of point cloud\cite{GFT, kipf2017semisupervised}, we propose to add the point clouds in the spectral domain and utilize the inverse transformation to decode the spectral features back to point-based object in the data domain.
Specifically,
given a point cloud $P$, we first take each point as a node and construct a K-NN graph on the whole point cloud by spatial distance, then compute its spectral graph Laplacian matrix $L_{n\times n}$ by $L = D - A$, where $D$ is the degree matrix and $A$ is the adjacency matrix. Next, we perform eigen-decomposition $L = Q\Lambda Q^T$ to acquire the orthonormal eigenvector matrix $Q$, which serves as the Graph Fourier Transform(GFT) basis. Therefore, the GFT coefficients $\phi_{GFT}(P) \in \mathbb{R}^{n\times 3}$, also called the spectral features, is then obtained by:
\begin{equation}
  \phi_{GFT}(P) = Q^TP.
\end{equation}
Here,
$\phi_{GFT}(P)$ performs as the spectral feature of the point cloud $P$ and reflects the geometry feature of $P$. It is more proper to mix the spectral feature of the point clouds, because in eigen-decomposition, we caculate ordered eigenvalues of $L$ as the diagonal matrix $\Lambda$. It makes sense to do linear addition on spectral features now and the general process is also clear now: For a point cloud $P$, we denote its spectral representation as $\widetilde{P}$, and apply spectral-aware Admix-based method on $\widetilde{P}$ to generate adversarial spectral feature $\widetilde{P^{adv}} = \widetilde{P} + \Delta$ by:
\begin{equation}
  \bar{g}_{t+1} = \frac{1}{m_1 \cdot m_2}\sum_{\widetilde{P}' \in S}^{} \sum_{i=0}^{m_1 - 1}\nabla _{\widetilde{P}^{adv}_t}J(\gamma_i\cdot(\widetilde{P}^{adv}_t+\eta_i \cdot \widetilde{P}'),y;\theta),
\end{equation}
At last, we revert $\widetilde{P^{adv}}$ to adversarial point cloud $P^{adv}$ in the data domain by using inverse GFT:
\begin{equation}
  P^{adv} = \phi_{IGFT}(\phi_{GFT}(P) + \Delta) = Q\widetilde{P^{adv}}.
\end{equation}

\subsection{Learning Weights for Better Guidance}
In order to guide the adversarial example to the decision boundary effectively and constrain the perturbation size, we need to apply weights on both the example mix step (i.e. $\gamma_i$ and $\eta_i$) and the gradient mix step in our spectral-aware Admix baseline. These learnable weights updated like $\widetilde{P^{adv}}$ can make better use of the spectral features we computed above in long iteration and guide $\widetilde{P^{adv}}$ to the decision boundary between other categories, which has strong transferability.

Naturally, we think that point clouds closer to the original point cloud should have larger weights, therefore we utilize the distance function to define the weights. Specficially, we use $w(\widetilde{P}, \widetilde{P}') = exp(-Dist(\widetilde{P}, \widetilde{P}'))$ rather than $w(\widetilde{P}, \widetilde{P}') = -Dist(\widetilde{P}, \widetilde{P}')$ to avoid scale problem and constrain the weights to $[0, 1]$. As for the distance, instead of using the general MSE, Chamfer, or Hausdorff distance of spatial positions, we attempt to use the distance of spectral features because it is already ordered and geometry-aware. Hence, we use weighted Euclidean distance as a part of mix weights to perform gradient descent method on it(in this part, we simply treat each column of $\widetilde{P}$ as a channel):
\begin{equation}
  Dist(\widetilde{P}, \widetilde{P}') = \sqrt{(\widetilde{P} - \widetilde{P}')^TM(\widetilde{P} - \widetilde{P}')},
\end{equation}
where $M_{n\times n}$ is a diagnoal positive matrix indicating the weights and we try to adjust it in a learnable manner. The Mahalanobis distance \cite{Mahalanobis1936OnTG} is a classic multivariate distance metric that measures the distance between a vector and a distribution. It has excellent applications in multivariate anomaly detection and classification on highly imbalanced datasets. It uses the inverse of the covariance matrix $\Sigma$ of the data $x$ to achieve scale-normalization among the components of $x$. However, due to $rank(AB)\le rank(A_{m\times n}) \le min(m, n)$, the covariance matrix $\Sigma_{n\times n}$ is not invertible. The number of points $n$ (e.g., 1024) is usually much bigger than the batch size $b$ (e.g., 64, 32 or maybe). On the other hand, updating a $1024\times1024$ matrix means lots of calculation cost and higher memory cost. Inspired by the above knowledge, we use the inverse of the variance matrix as the initial $M_0$ and we constrain the matrix $M$ to be diagonal all the time. $M$ is already semi-positive-definite now and we can constrain it to be positive-definite simply by constraining the diagnoal elements of $M$ positive. $M$ updated in this manner also have more numerical stability than using Mahalanobis-like distance and updating a $n\times n$ weight matrix.

As for the initialization of $M$, we use the inverse of the variance, and slightly changed it to prevent value error:
\begin{equation}
M_0 = Diag(1/(Var + \epsilon))
\end{equation}

Where $\epsilon$ is a small positive number and the operations are element-wised. By using some mathematical properties of matrices, we derive our weights with the learning of $M$, which make better use of the geometry feature of the point clouds and guide the adversarial examples to the decision boundaries better.

\subsection{Fixed weights for better imperceptibility}

Learning weights in the sample space can guide our adversarial examples better to the decision boundaries, but we also need to maintain the Geometry structure and the imperceptibility of the adversarial samples. Noticing that the lower components of the spectral feature represents the shape and the higher components represents the details\cite{hu2022exploring}. The energy of the spectral feature concentrates on the top-32 dimensions, so we use different weights in the lower frequency and the higher frequency:
\begin{equation}
M_s = \alpha_{low}\mathbf{1}_{low}+\alpha_{high}\mathbf{1}_{high}
\end{equation}
Where $0<\alpha_{high}<\alpha_{low}<1$ are the mixing weights. $\mathbf{1}_{low}$ values 1 at top-32 elements and 0 at other elements, $\mathbf{1}_{high} = I - \mathbf{1}_{low}$. Next, we mix the point clouds with the two weights we calculated before to make the adversarial point cloud close to the original point cloud in the lower spectral components, thus maintaining the the Geometry structure and the imperceptibility:
\begin{equation}
  \bar{g}_{t+1} = \frac{1}{m_1 \cdot m_2}\sum_{\widetilde{P}' \in S}^{} \sum_{i=0}^{m_1 - 1}\nabla _{\widetilde{P}^{adv}_t}J(\gamma_i  M_s\widetilde{P}^{adv}_t+\gamma_i \eta_i (I - M_s) \widetilde{P}',y;\theta),
\end{equation}

\subsection{Exploration of the Mixing Path}
In addition to the mixed weights we analysed above, we also need to select the augmentation paths\cite{zhang2023improving} to direct our adversarial examples to the decision boundaries of other categories better. Like data augmentation, we copy and adjust the original examples to generate a more transferable example in a learnable way. The better we select the path, the better transferability we can acquire. With larger batch size, for each point cloud in the batch, we can find many point clouds from other categories, which means many augmentation paths, but the amount is not fixed for every point cloud in the batch. What's more, using all point clouds from other categories means higher time cost and some candidates may do harm to the transferability, so we need to select a mixing path to better steer the adversarial point cloud to the decision boundary. A natural idea is to run a few iterations, use some metrics to select better augmentation path, and run more iterations through this augmentation path. Adversarial loss is a classical metric to select the path, but it doesn't fully reflect the trend of loss at the last iteration. Inspired by GRA\cite{10377670}, we use the cosine similarity to select the paths. Define that $cos(g, g') = \frac{gg'}{\left | g \right | \left | g' \right | } $ is the cosine similarity of the gradients:

\begin{equation}
  \bar{g}_{t+1}^j = \frac{1}{m_1}\sum_{i=0}^{m_1 - 1}\nabla _{\widetilde{P}^{adv}_t}J(\gamma_i\cdot(\widetilde{P}^{adv}_t+\eta_i \cdot \widetilde{P}'),y;\theta),
\end{equation}

\begin{equation}
  g_{t+1}^{adv} = \nabla _{\widetilde{P}^{adv}_t}J(\widetilde{P}^{adv}_t, y; \theta).
\end{equation}

In specific, we first select many examples from other categories randomly and run a few iteration as Equation (3). At the t-th epoch, we calculate the average gradient $\bar{g}_{t+1}^j$ for each candidate $\widetilde{P}'$ and the gradient $g_{t+1}^{adv}$ of the adversarial example $\widetilde{P}^{adv}_t$. Next, we calculate the cosine similarity $cos(g_{t+1}^{adv}, \bar{g}_{t+1}^j)$ and select the mixing paths. More cosine similarity means better direction and we need to turn our direction towards it. Finally, we use weighted Admix on the selected paths to steer the adversarial example to the decision boundaries. By selecting better paths, we can generate better adversarial examples with less perturbation.

\subsection{Generating Adversarial Example}

Recall that the goal of our task is to generate adversarial example $P^{adv}$ for each point cloud with the original examples in the batch and the gradients of the known surrogate model $F_\psi$. We want the examples we generate can cheat any unknown target classification model $F_\theta$. We use the clip function in both spectral domain and spacial domain to constrain the distance of adversarial example to the original example and we use the mixed loss below to perform gradient descent on both adversarial example $\widetilde{P^{adv}}$ and learnable weights $M$:
\begin{equation}
  L_{adv}(P', y) = max(F(P', y) - \max_{y'\neq y}F(P', y'), 0),
\end{equation}
\begin{equation}
  Loss(P, P', y) = L_{adv}(P', y) + \lambda_1 D_{n}(P, P') + \lambda_2 D_{c}(P, P') + \lambda_3 D_{h}(P, P')
\label{loss}
\end{equation}
Where $P$ is the original point cloud and $P'$ is the adversarial point cloud, $y$ is the ground truth label for $P$, and $F$ is the classifier. $D_{n}, D_{c}, D_{h}$ are MSE, Chamfer loss and Hausdorff loss and $\lambda$ are their weights. Adversarial examples guided by such loss balanced the logit of $y$ and the distance from $P'$ to $P$. By adjusting $\lambda$, we can generate both higher imperceptibility examples and higher ASR examples. So far, we complete our generating method, and the optimization algorithm is shown below, noting that we reuse $Mix(\cdot)$ as weighted-Admix method:
\begin{algorithm}
\caption{spectral-awared Admix}\label{algorithm1}
\KwIn{target point cloud $P$, mix point clouds $P_1, \cdots, P_n$, initial perturbation $\Delta$, step $k$, initial mix weights matrix $M$, spectral mix mask $M_s$, label $y$}
\KwOut{adversarial point cloud $P_k^{adv}$, optimized weights $M$, gradient $\bar{g}_{k}^j$ and $g_{k}^{adv}$}
$\mathbf{def} \ f(P_1, P_2; M) = exp(-\sqrt{(\widetilde{P_1} - \widetilde{P_2})^TM(\widetilde{P_1} - \widetilde{P_2})})$\;
$P_0^{adv} = P$\;
$t = 0$\;
Initialize Admix lower bound $b_l$ and upper bound $b_u$\;
\While{$t < k$}{$\beta_i = b_l+\frac{i}{m-1}(b_u-b_l)$\;$\widetilde{P_t^{i, j}} = \beta_iM_s\widetilde{P_t^{adv}}+(1-\beta_i)(I-M_s)f(P, P_j;M)\widetilde{P_j}$\;$g_{t+1}^\Delta=\frac{1}{mn}\sum_{j=1}^{n}\sum_{i=0}^{m-1}f(P, P_j;M)\partial_\Delta(Loss(P, P_t^{i, j}, y))$\;$\Delta = clip_\Delta(\Delta - lr \cdot g_{t+1}^\Delta)$\;$g_{t+1}^M=\frac{1}{mn}\sum_{j=1}^{n}\sum_{i=0}^{m-1}\partial_M(Loss(P, P_t^{i, j}, y))$\;$M = clip_M(M - lr\cdot g_{t+1}^M)$\;$\widetilde{P_{t+1}^{adv}} = \widetilde{P} + \Delta$\;$t = t+1$\;}
$g_k^{adv}= \partial_{\widetilde{P}^{adv}_t}(Loss(P, P^{adv}_k, y))$\;$\bar{g}_{k}^j=\frac{1}{m}\sum_{i=0}^{m - 1}\partial_{\widetilde{P}^{adv}_t}(Loss(P, P_k^{i, j}, y))$
\end{algorithm}
\begin{algorithm}
\caption{Geometry-aware point cloud attack}\label{algorithm2}
\KwIn{a batch of point clouds $P_1, \cdots, P_b$ with label $l_1, \cdots, l_b$}
\KwOut{adversarial point clouds $P_1^{adv}, \cdots, P_b^{adv}$}
Initialize $M_0 = Diag(1/(Var + \epsilon))$ and $M_s$\;
\For{$t = 1\ \mathbf{to}\ b$}{Find $n'$ point clouds $P_{t'_1}, \cdots, P_{t'_{n'}}$ with $l_{t'_j} \neq l_t$\;Initialize perturbation $\Delta$ \;
$\_, M, g_{k}^{adv}, \bar{g}_{k}^j = Mix(P_t, \Delta, k', M, M_s, y; P_{t'_j})$\;Search $n$ paths with largest cosine similarity\;
$P_t^{adv}, M, \_, \_ = Mix(P_t, \Delta, k, M, M_s, y; P_{t_j})$\;}
\end{algorithm}

\begin{table*}[t!]
\begin{tabular}{|c|ccc|ccc|ccc|ccc|}
\hline
\multirow{2}{*}{Attack} & \multicolumn{3}{c|}{PointNet}        & \multicolumn{3}{c|}{PointNet++}           & \multicolumn{3}{c|}{PointConv}               & \multicolumn{3}{c|}{DGCNN}          \\ \cline{2-13} 
                        & $D_h$ & $D_c$        & $D_{norm}$    & $D_h$        & $D_c$ & $D_{norm}$    & $D_h$        & $D_c$        & $D_{norm}$    & $D_h$        & $D_c$        & $D_{norm}$      \\ \hline
FGSM & 0.1853 & 0.1326 & 0.7936 & 0.2275 & 0.1682 & 0.8357 & 0.2379 & 0.1544 & 0.8322 & 0.2506 & 0.1890 & 0.8549\\
3D-Adv & 0.0105 & \textbf{0.0003} & 0.3032 & 0.0381 & \textbf{0.0005} & 0.3248 & 0.0405 & 0.0006 & 0.3218 & 0.0475 & \textbf{0.0005} & 0.3326\\
GeoA & 0.0175 & 0.0064 & 0.4385 & 0.0357 & 0.0198 & 0.4772 & 0.0383 & 0.0185 & 0.4837 & 0.0402 & 0.0176 & 0.4933 \\
AdvPC & 0.0343 & 0.0528 & 0.4328 & 0.0429 & 0.0781 & 0.4557 & 0.0471 & 0.0697 & 0.5173 & 0.0148 & 0.0844 & 0.3975\\
SS-Attack & 0.0129 & 0.0028 & 0.2936 & 0.0154 &  0.0037 & 0.3142 & 0.0218 & 0.0024 & 0.2592 & 0.0097 & 0.0061 & 0.3006 \\ \hline
Ours-B & 0.0032 & 0.0015 & 0.1630 & 0.0054 & 0.0011 & 0.1367 & 0.0040 & 0.0008 & 0.1475 & 0.0035 & 0.0023 & 0.2194\\
Ours-F & \textbf{0.0028} & 0.0009 & \textbf{0.1583} & \textbf{0.0039} & 0.0007 & \textbf{0.1395} & \textbf{0.0031} & \textbf{0.0005} & \textbf{0.1267} & \textbf{0.0024} & 0.0013 & \textbf{0.1725}\\
\hline
\end{tabular}
\caption{Average distances of adversarial examples from original examples. $D_h$ means Hausdorff distance, $D_c$ means Chamfer distance, $D_{norm}$ means Mean Square Error.}
\label{result}
\end{table*}

\begin{table*}[t!]
\begin{tabular}{|c|c|cccc|}
\hline
Model                      & Attack & PointNet        & PointNet++          & PointConv        & DGCNN               \\ \hline
\multirow{7}{*}{PointNet}  & FGSM & 100.0\%* & 3.9\% & 3.8\% & 0.6\% \\
~ & 3D-Adv & 100.0\%* & 5.3\% & 2.3\% & 3.8\% \\
~ & GeoA & 100.0\%* & 11.6\% & 7.4\% & 2.6\% \\
~ & AdvPC & 100.0\%* & 30.4\% & 13.2\% & 14.6\%  \\
~ & SS-Attack & 99.9\%* & 58.6\% & 37.9\% & 30.0\% \\
~ & Ours-B & 100.0\%* & 71.0\% & 45.2\% & 45.9\% \\
~& Ours-F & 100.0\%* & 82.8\% & 61.9\% & 64.3\% \\ \hline
\multirow{7}{*}{PointNet++}  & FGSM & 3.2\% & 100.0\%* & 4.4\% & 5.6\% \\
~ & 3D-Adv & 2.09\% & 78.5\%* & 4.95\% & 6.27\% \\
~ & GeoA & 9.8\% & 100.0\%* & 13.5\% & 19.8\% \\
~ & AdvPC & 5.08\% & 99.5\%* & 27.5\% & 17.5\% \\
~ & SS-Attack & 7.57\% & 97.2\%* & 46.1\% & 31.1\%  \\
~ & Ours-B & 32.8\% & 100.0\%* & 58.9\% & 52.7\%  \\
~& Ours-F & 56.9\% & 100.0\%* & 71.6\% & 69.0\% \\ \hline
\multirow{7}{*}{PointConv}  & FGSM & 4.8\% & 7.9\% & 99.8\%* & 6.5\% \\
~ & 3D-Adv & 1.81\% & 7.87\% & 84.3\%* & 4.47\% \\
~ & GeoA & 10.59\% & 32.96\% & 100.0\%* & 19.28\% \\
~ & AdvPC & 4.62\% & 35.3\% & 99.4\%* & 17.7\% \\
~ & SS-Attack & 5.49\% & 42.8\% & 98.5\%* & 21.0\% \\
~ & Ours-B & 34.62\% & 59.76\% & 100.0\%* & 46.15\%\\
~& Ours-F & 51.08\% & 68.31\% & 100.0\%* & 60.44\%\\ \hline
\multirow{7}{*}{DGCNN}  & FGSM & 3.6\% & 7.2\% & 9.6\% & 100.0\%* \\
~ & 3D-Adv & 0.99\% & 5.84\% & 4.76\% & 100.0\%* \\
~ & GeoA & 12.5\% & 24.2\% & 28.9\% & 100.0\%* \\
~ & AdvPC & 6.94\% & 60.1\% & 43.5\% & 93.7\%* \\
~ & SS-Attack & 13.9\% & 68.2\% & 60.0\% & 100.0\%* \\
~ & Ours-B & 38.4\% & 76.2\% & 51.6\% & 100.0\%* \\
~& Ours-F & 61.5\% & 79.3\% & 68.4\% & 100.0\%* \\ \hline

\end{tabular}
\caption{Transfer-based Attacks on the ModelNet40 dataset, * means white box attack, and others mean black box attack without query.}
\label{blackbox}
\end{table*}

\begin{table*}[t!]
\begin{tabular}{|c|c|cccc|}
\hline
Defense                      & Attack & PointNet        & PointNet++          & PointConv        & DGCNN                   \\ \hline
\multirow{5}{*}{SRS}  & 3D-Adv & 36.9\% & 33.8\% & 14.5\% & 20.7\% \\
~ & AdvPC & 89.4\% & 63.3\% & 29.0\% & 33.6\%  \\
~ & SS-Attack & 96.6\% & 51.7\% & 24.3\% & 31.5\% \\
~ & Ours-B & 98.3\% &  60.8\% & 39.7\% & 34.8\%  \\
~& Ours-F & 99.6\% &  72.4\% & 49.6\% & 52.5\% \\ \hline
\multirow{5}{*}{SOR} & 3D-Adv & 18.3\% & 4.1\% & 3.2\% & 5.7\%\\
~ & AdvPC & 53.2\% & 16.3\% & 6.4\% & 10.5\% \\
~ & SS-Attack & 73.4\% & 9.3\% & 5.2\% & 9.5\%\\
~ & Ours-B & 94.2\% & 48.7\% & 32.5\% & 26.4\% \\
~& Ours-F & 96.1\% & 63.9\%  &  39.6\% & 38.7\% \\ \hline
\multirow{5}{*}{DUP-Net} & 3D-Adv & 9.9\% & 5.9\% & 7.8\% & 8.7\%\\
~ & AdvPC & 23.8\% & 16.8\% & 12.0\% & 14.7\% \\
~ & SS-Attack & 47.3\% & 9.8\% & 8.6\% & 11.7\%\\
~ & Ours-B & 74.1\% & 32.6\% & 23.8\% & 22.1\%  \\
~& Ours-F & 78.4\% & 47.5\% & 30.7\% & 28.2\% \\ \hline
\multirow{5}{*}{ConvNet-Opt} & 3D-Adv & 6.2\% & 11.2\% & 9.6\% & 11.0\%\\
~ & AdvPC & 9.9\% & 13.8\% & 11.4\% & 12.5\% \\
~ & SS-Attack & 17.0\% & 12.1\% & 9.6 \% & 11.5\% \\
~ & Ours-B & 52.9\% & 21.8\% & 14.7\% & 19.4\% \\
~& Ours-F & 59.8\% & 26.3\% & 18.0\% & 17.6\% \\ \hline
\multirow{5}{*}{Onet-Remesh}  & 3D-Adv & 11.5\% & 24.2\% & 20.1\% & 24.6\%\\
~ & AdvPC & 12.7\% & 27.4\% & 23.8\% & 26.4\%  \\
~ & SS-Attack & 14.0\% & 25.5\% & 21.8\% & 26.9\%\\
~ & Ours-B & 62.0\% & 27.3\% & 22.9\% & 18.7\% \\
~& Ours-F & 67.5\% & 35.8\% & 26.4\% & 20.1\%  \\ \hline
\multirow{5}{*}{ONet-Opt}  & 3D-Adv & 6.6\% & 13.9\% & 13.1\% & 13.0\% \\
~ & AdvPC & 8.0\% & 17.9\% & 15.1\% & 13.6\% \\
~ & SS-Attack & 11.7\% & 15.7\% & 13.5\% & 14.3\% \\
~ & Ours-B & 39.8\% & 17.9\% & 16.7\% & 20.9\% \\
~& Ours-F & 46.5\% & 20.4\% & 19.2\% & 18.3\% \\ \hline
\end{tabular}
\caption{Transfer-based Attacks on the ModelNet40 dataset on PointNet model against various defense strategies.}
\label{defense}
\end{table*}

\section{Experiments}
\subsection{Dataset and 3D models}
\noindent \textbf{Dataset.} Just like previous point cloud classification and point cloud attack works, We use ModelNet40\cite{wu20153d} in our experiments to evaluate our attack performance. This dataset contains 12,311 CAD models from 40 object categories (such as airplanes, desks, chairs and lights), 9,843 as training data and 2,468 as testing data. We use the ModelNet40 normal-resampled dataset, in which 1,024 points are normally sampled from the surface of each object and the points are scaled in a unit ball. For the adversarial point cloud attacks, we randomly select a number of instances in the ModelNet40 testing set, which can be well classified by the classifiers of interest.

\noindent \textbf{3D Models.} We use four point cloud classification models: PointNet, PointNet++\cite{qi2017pointnet}, PointConv\cite{wu2020pointconv} and DGCNN\cite{wang2019dynamic}. For each classification model, we use it as the surrogate model and use the adversarial samples generated on it to attack all four classification models.

\subsection{Implementation Details}
\noindent \textbf{Attack Setup.} 
To generate the adversarial examples, we adopt the Adam optimizer to optimize the objective of our proposed attack. We use a fixed learning schedule of 500 iterations, where the learning rate and momentum are set as 0.01 and 0.9, respectively.
In the loss function Eq.\ref{loss}, we set $\lambda_1, \lambda_2, \lambda_3$ to 0.5, 20, 50. In the spectral feature extraction, we use $k = 10$ to build a K-NN graph. In the adjusted Admix (Algorithm. \ref{algorithm1}), we set $b_l = 0.1, b_h = 0.9, \alpha_{low} = 0.9, \alpha_{high} = 0.25$, and $m = 20, n = 9$. All experiments are implemented on a single NVIDIA RTX 2080Ti GPU.

\noindent \textbf{Evaluation Metrics.} We use three classic point cloud distances: L2-norm distance ($D_{norm}$), Chamfer distance ($D_c$), and Hausdorff Distance($D_h$) as our metrics to evaluate the perturbation size. On the other hand, we evaluate the attack performance simply by comparing the attack success rate. Noting that we did weighted addition on these three distances in our loss function with the adversarial loss to guide the adversarial examples to be more imperceptible.


\subsection{Evaluation on our attack}
\noindent \textbf{Comparison on the perturbation size.} We compare the performance with several existing white box attacks and black box attacks without query, and the results are shown in Table \ref{result}. ``Ours-B" denotes our method solely with spectral-aware admix, ``Ours-F" denotes our full attack method.
From this table, our full attack generates adversarial point clouds with almost the lowest perturbation sizes in all evaluation metrics on five attack models. Note that, 3D-ADV has the best Chamfer distance as the version of 3D-ADV that we compare to generates and places a set of independent points close to the original object, without perturbing existing points in the point cloud. However, it induces much larger distortions than ours on the other four metrics. In particular, we require the lowest perturbation size in terms of the geometric distortion metrics, which indicates that our generated adversarial point clouds reveal the geometric fidelity best. Compared to the GeoA which optimizes a specific curvature loss with xyz-directional perturbations to keep the original geometry information, our attack constrains the perturbation along the best optimization path to directly keep the position dependency among neighboring points, thus promoting the consistency of local curvatures and achieving smaller distances. Besides, for each attack method, it takes larger perturbation sizes to successfully attack PointNet++, PointConv than to attack PointNet and DGCNN, which indicates that PointNet++, PointConv are harder to attack.

\noindent \textbf{Comparison on the transferability.}
We perform extensive experiments to prove that our attacks are more transferable to different types of 3D point cloud classifiers. We compared our attacks with the adversarial examples generated by FGSM\cite{goodfellow2014explaining}, 3D-Adv\cite{9586275}, GeoA\cite{wen2020geometry}, AdvPC\cite{hamdi2020advpc} and SS-Attack\cite{long2022frequency}. We implement the untargeted attack, making the perturbed point cloud to be classified as the ground truth label by the target model, not constraining it to be the fixed label we choose. In Table.\ref{blackbox}, we compared the attack success rates of the attack methods, * means white box attack and others mean black box attack without query. The rows are the surrogate models and the columns are the target models. Our attack methods have a significant improvement of transferbility, we can get better attack success rate with almost least perturbation even without the path selection module. Besides, the white box attack success rate remains 100\%. The black box attack success rates basically increased by at least 10\%. On average, the black box attack success rate increased by 11.8\% on PointNet, 19.1\% on PointNet++, 22\% on PointConv and 8\% on DGCNN, one failure in DGCNN also implies that it is a harder classification model to attack. 

The effectiveness of the path selection module is also clear. The black box attack success rates further increased by at least 3\% than our attack without path selection and the white box attack remains 100\%. On average, the black box attack success rate increased by 15.6\% on PointNet, 17.7\% on PointNet++, 13.3\% on PointConv and 14.3\% on DGCNN compared with our attack method without path selection strategy. Our attack with path selection achieves the best among all the existing attack methods we select.


\noindent \textbf{Resistance to defenses.}We further tests our attack method's performance under various kinds of defense methods. We selected Simple Random Sampling (SRS)\cite{yang2021adversarial}, Statistical Outlier Removal (SOR)\cite{yu2018multi}, DUP-Net\cite{zhou2019dup} and three IF-Defense\cite{wu2020if} methods ConvNet-Opt, Onet-Remesh, Onet-Opt. We compared the attack success rate under these defense methods using PointNet as the surrogate model. In table\ref{defense}, we can see that our methods under most defenses are still considerably better than other attack methods. In the white box attack, our attack methods remains very high attack success rate under SRS and SOR, and our methods are higher than other attack methods at least 20\% under other defense methods. That is because our methods generate less outliers and maintain local point distribution. As for the black box attack, our method without path selection decreased on average by 8.9\% under SRS, 18.2\% under SOR, 27.9\% under DUP-Net, 35.4\% under ConvNet-Opt, 31.1\% under Onet-Remesh and 35.6\% under Onet-Opt. Our method with path selection decreased on average by 11.5\% under SRS, 22.3\% under SOR, 34.2\% under DUP-Net, 49\% under ConvNet-Opt, 42.2\% under Onet-Remesh and 50.4\% under Onet-Opt. It shows that our path selection strategy is effective against the defense methods, and IF-defenses have better defense performance by capturing the clean shape and optimizing clean and complete point cloud under geometry- and distribution- aware constraints.

\section{Conclusion}
In this paper, we propose a novel Spectral-aware Admix with Augmented Optimization method (SAAO) to improve the transferability of 3D point cloud attack.
Inspired by the 2D admix strategy, we design a new spectral-aware admix by incorporating Graph Fourier Transform to analysis the frequency-reflected geometrics for preserving the object shape during the perturbation optimization. An augmentation path strategy is also introduced to help find the best optimization patch for improving the transferability. 
Experiments demonstrate that our proposed attack can achieve better transferability than existing methods, generate more imperceptible adversarial examples, and break through several defense methods notably.


\newpage

\bibliographystyle{ACM-Reference-Format}
\bibliography{sample-base}


\end{document}